\documentclass[letterpaper]{article} 
\usepackage{aaai23}  
\usepackage{times}  
\usepackage{helvet}  
\usepackage{courier}  
\usepackage[hyphens]{url}  
\usepackage{graphicx} 
\usepackage{natbib}  
\usepackage{caption} 
\usepackage{algorithm}
\usepackage{algorithmic}
\usepackage[switch]{lineno}
\usepackage{amssymb}
\usepackage{multirow}
\usepackage{multicol}
\usepackage{array}
\usepackage{amsmath}
\usepackage[dvipsnames, svgnames, x11names]{xcolor}

\usepackage{newfloat}
\usepackage{listings}
\DeclareCaptionStyle{ruled}{labelfont=normalfont,labelsep=colon,strut=off} 
\floatstyle{ruled}
\newfloat{listing}{tb}{lst}{}
\floatname{listing}{Listing}
\title{Chinese Character Recognition with Radical-Structured Stroke Trees}
\author{
    Haiyang Yu, Jingye Chen, Bin Li\thanks{Corresponding author}, Xiangyang Xue
}
\affiliations{
Shanghai Key Laboratory of Intelligent Information Processing\\
 School of Computer Science, Fudan University \\
 \{hyyu20, jingyechen19, libin, xyxue\}@fudan.edu.cn
}

\usepackage{bibentry}

\begin{document}
\maketitle

\begin{abstract}
The flourishing blossom of deep learning has witnessed the rapid development of Chinese character recognition. However, it remains a great challenge that the characters for testing may have different distributions from those of the training dataset. Existing methods based on a single-level representation (character-level, radical-level, or stroke-level) may be either too sensitive to distribution changes (\textit{e.g.}, induced by blurring, occlusion, and zero-shot problems) or too tolerant to one-to-many ambiguities. In this paper, we represent each Chinese character as a stroke tree, which is organized according to its radical structures, to fully exploit the merits of both radical and stroke levels in a decent way. We propose a two-stage decomposition framework, where a Feature-to-Radical Decoder perceives radical structures and radical regions, and a Radical-to-Stroke Decoder further predicts the stroke sequences according to the features of radical regions. The generated radical structures and stroke sequences are encoded as a \textbf{R}adical-\textbf{S}tructured \textbf{S}troke \textbf{T}ree (RSST), which is fed to a Tree-to-Character Translator based on the proposed Weighted Edit Distance to match the closest candidate character in the RSST lexicon. Our extensive experimental results demonstrate that the proposed method outperforms the state-of-the-art single-level methods by increasing margins as the distribution difference becomes more severe in the blurring, occlusion, and zero-shot scenarios, which indeed validates the robustness of the proposed method.
\end{abstract}

\begin{figure}[t]
    \centering
    \includegraphics[width=0.47\textwidth]{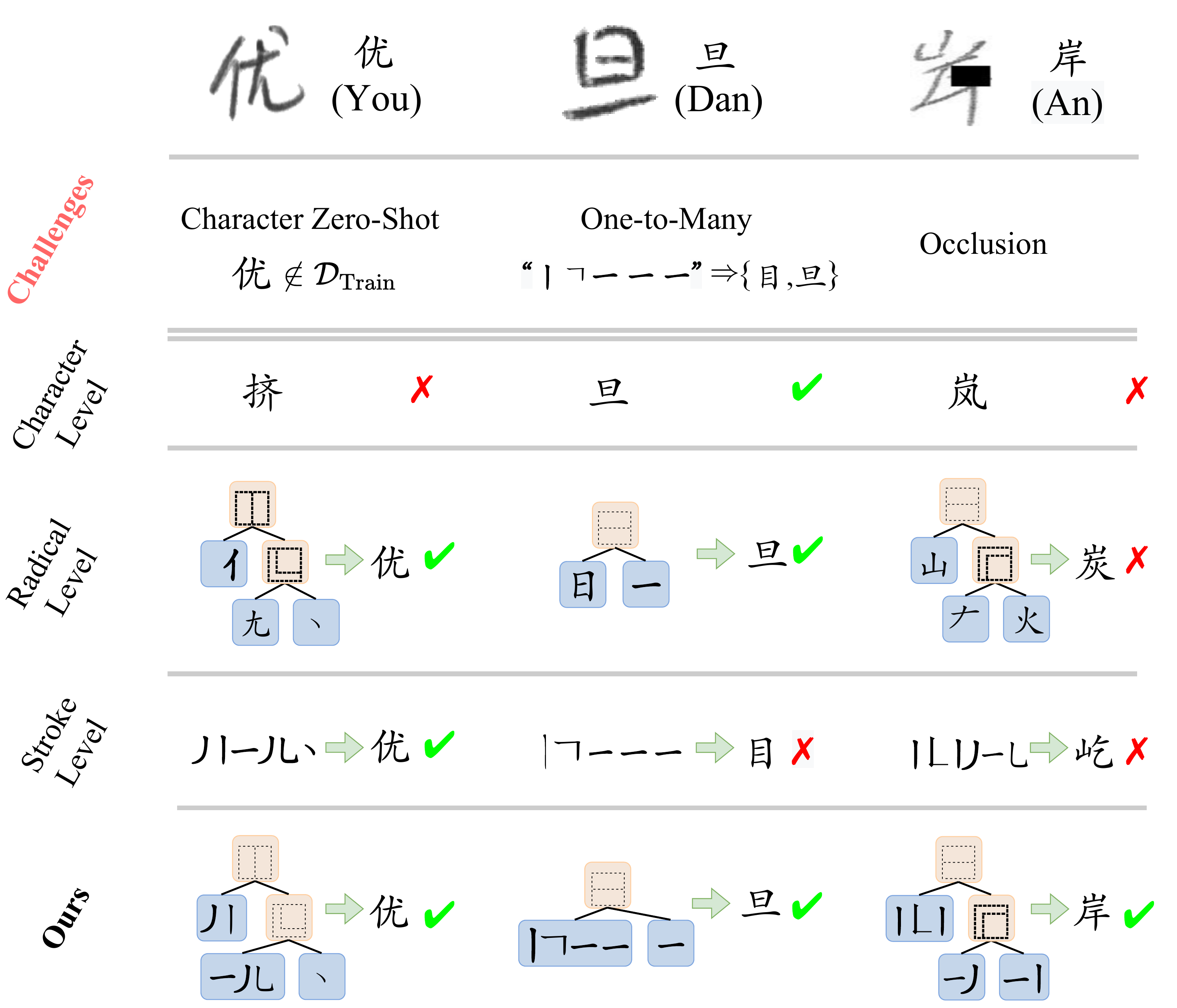}
    \caption{The proposed RSST representation can address the challenges that bring difficulties to  existing single-level methods. The ground truth and Pinyin are on the right side of each character image. $\mathcal{D}_{\text{Train}}$ denotes the training dataset and $\Rightarrow$ denotes the stroke-to-character correspondence.}
    \label{fig:intro}
\end{figure}

\section{Introduction}
As a language used by a huge population of 1.31 billion people, Chinese character recognition (CCR) has attracted extensive research interests \cite{pal2012off,yang2015chinese,meng2001generating,xiaobo2003vehicle,Zu_mm,chen2021benchmarking}. Although CCR has been explored for more than thirty years, the existing methods are vulnerable to some challenging situations, especially when the distribution of characters for testing differs from that for training. Concretely, the discrepancies in the categorical distribution (\textit{e.g.}, characters to be tested are absent from the training dataset, namely zero-shot problems) or the visual distribution (\textit{e.g.}, blurring or occlusions in real-world scene images) still brings difficulties to the existing methods.

Existing CCR methods can be categorized into three approaches in terms of the recognized components, including \textit{character-based}, \textit{radical-based}, and \textit{stroke-based} approaches. The character-based approach \cite{xiao2017building,zhang2017online,wang2021joint} usually perceives each character as a unique category and captures the global features such as contours to obtain the character predictions. However, it is sensitive to the categorical distribution changes, especially when the tested character does not appear in the training dataset, namely the character zero-shot problem (see the example ``You'' in Fig. \ref{fig:intro}). To tackle the character zero-shot problem, the radical-based approach \cite{wang2018denseran,wang2019radical} decomposes characters at the radical level and represents each Chinese character as a tree of radicals organized according to its radical structures. However, although it can alleviate the character zero-shot problem to some extent, another conundrum called the radical zero-shot problem may arise if some uncommon radicals are absent from the training dataset. Moreover, the radical-based approach is sensitive to the visual distribution changes induced by blurring and occlusions (see the example ``An'' in Fig. \ref{fig:intro}). Further, the stroke-based approach \cite{chen2021zero} is proposed to decompose characters into stroke sequences (\textit{i.e.}, the atomic units of Chinese characters), which can fundamentally overcome the categorical distribution changes. However, this approach leads to an undesirable one-to-many problem as one stroke sequence may correspond to more than one character (see the example ``Dan'' in Fig. \ref{fig:intro}). Therefore, none of the aforementioned single-level approaches can tackle all these challenges by itself. In this paper, we seek to answer this question: \textit{Is there a more decent way to represent Chinese characters for tackling CCR?}

To answer this question, we view each Chinese character as a \textbf{R}adical-\textbf{S}tructured \textbf{S}troke \textbf{T}ree (RSST) to fully exploit the merits of both radical-level and stroke-level representations. Specifically, we propose a framework consisting of a Feature-to-Radical Decoder (FRD) and a Radical-to-Stroke Decoder (RSD). The former decoder predicts the radical structures of the input character image and perceives the radical regions; the latter takes the features of radial regions as input to predict corresponding radical sequences. The intermediate outputs of FRD and RSD are integrated into a RSST to represent each Chinese character. Considering the stroke tree may not precisely match one specific character, in the Tree-to-Character Translator (TCT), we design the Weighted Edit Distance as the metric that takes both the radical structures and strokes into account. The extensive experimental results show that our method outperforms the existing single-level methods by increasing margins as the distribution difference becomes more severe in the blurring, occlusion, and zero-shot scenarios, which indeed validates the robustness of the proposed RSST representation. In summary, the contributions can be listed as follows:
\begin{itemize}
    \item We represent each Chinese character as a RSST to exploit the merits at the radical and stroke levels for tackling CCR.
    \item We propose a hierarchical decomposition framework, where a Feature-to-Radical Decoder predicts the radical structures and the locations of radicals, and a Radical-to-Stroke Decoder predicts the strokes of each radical.
    \item To alleviate the one-to-many problem, we propose the Weighted Edit Distance taking both radical structures and strokes into account to search for the most reasonable candidate in the pre-defined RSST lexicon.
    \item Our method outperforms the SOTA single-level methods when the distribution changes greatly during testing, which further validates the superiority of our method.
\end{itemize}

\begin{figure}[t]
    \centering
    \includegraphics[width=0.476\textwidth]{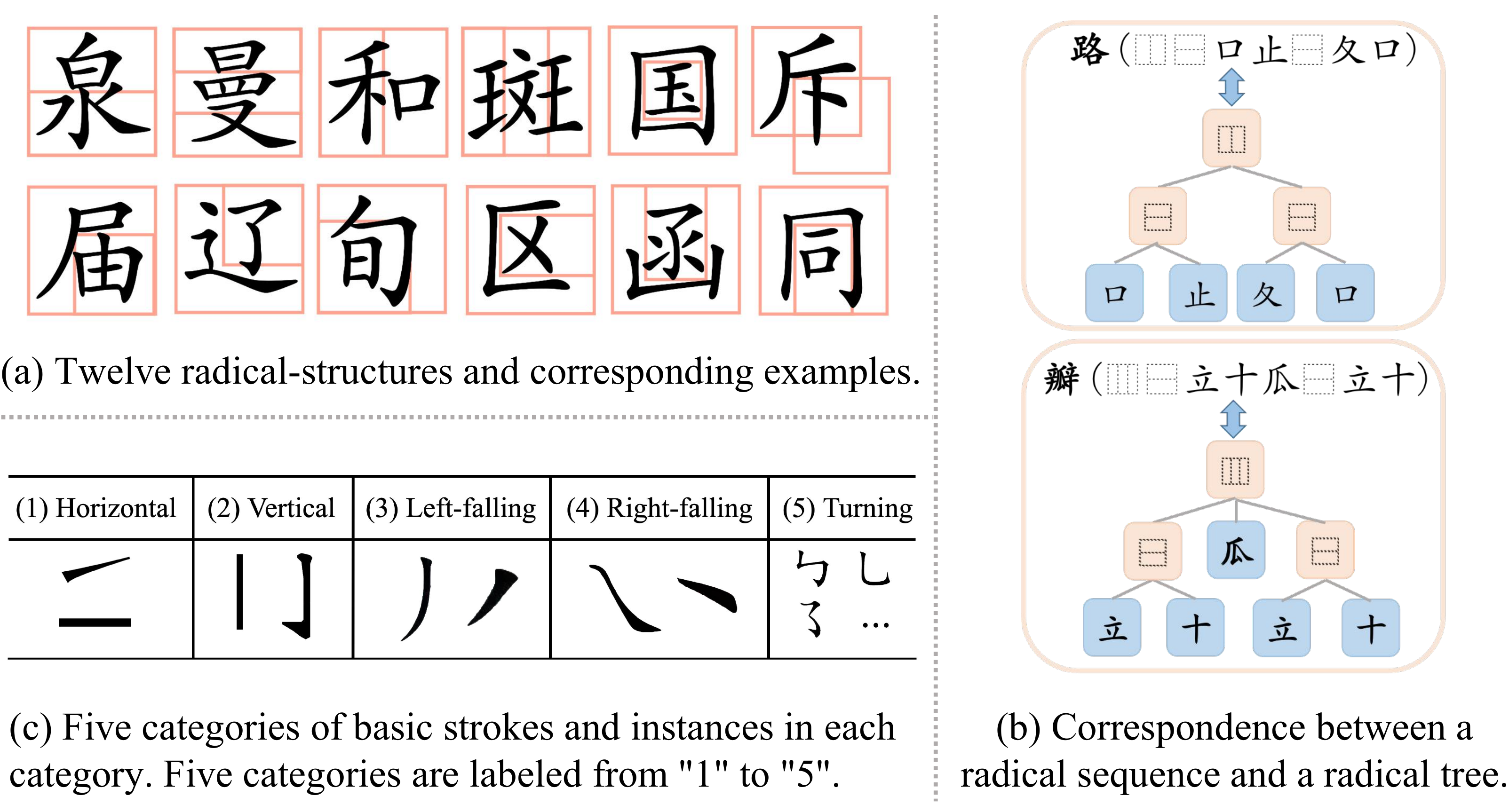}
    \caption{The preliminary knowledge of decomposing Chinese characters at the radical level and stroke level.}
    \label{fig:structure_and_stroke}
\end{figure}

\begin{figure*}[t]
    \centering
    \includegraphics[width=0.88\textwidth]{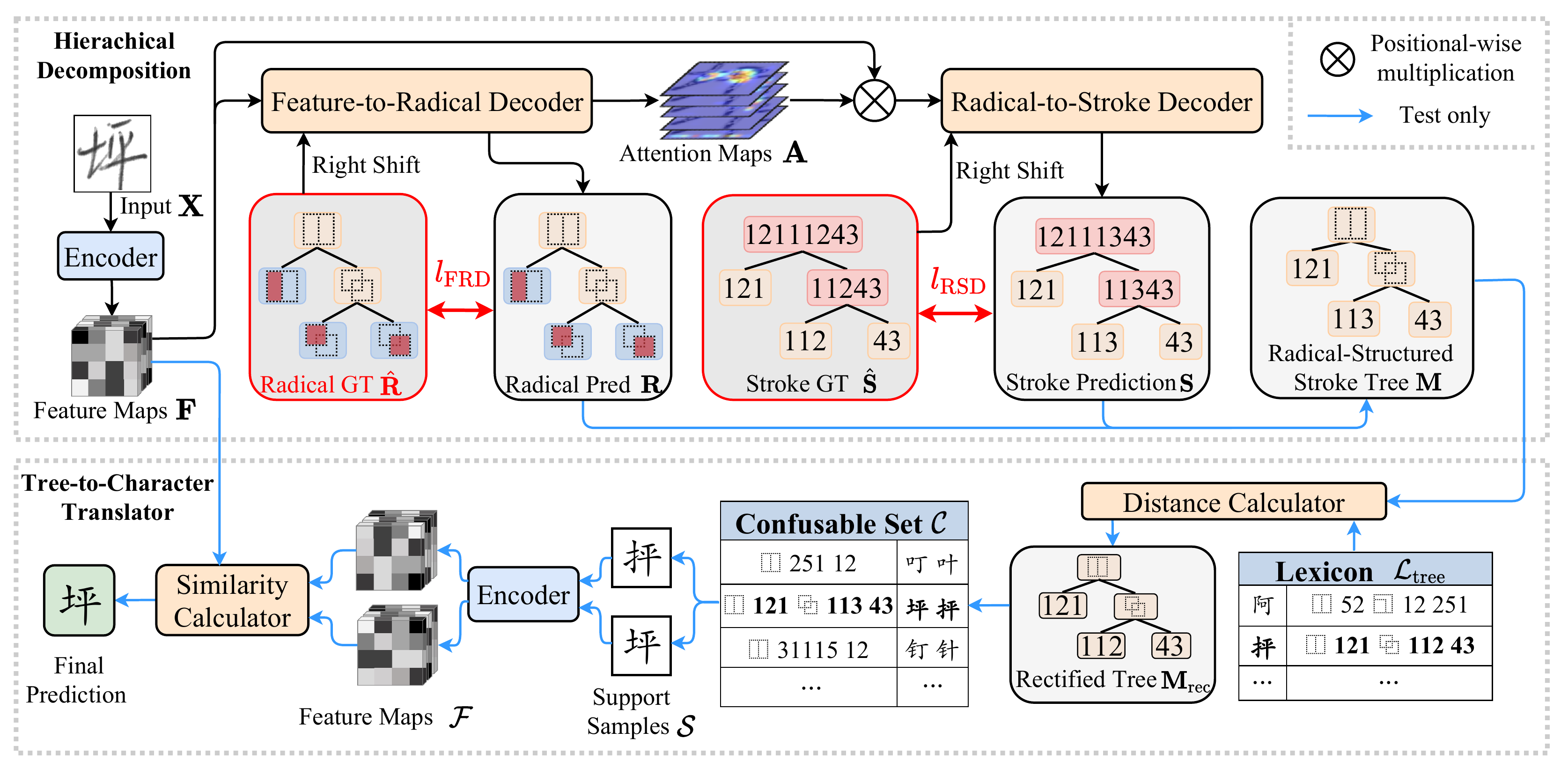}
    \caption{The overall architecture consists of a character decomposition framework. Radical structures in radical-level prediction $\mathbf{R}$ and strokes in stroke-level prediction $\mathbf{S}$ are combined into the RSST $\mathbf{M}$.}
    \label{fig:architecture_overall}
\end{figure*}

\section{Preliminaries}
\subsection{Chinese Character Decomposition}
According to the national standard GB18030-2005, there are 70,244 characters in total. Among them, 3,755 characters are commonly-used Level-1 characters. Each Chinese character can be decomposed into a radial or stroke sequence in a specific order. At the radical level, there are 12 radical structures (see Fig. \ref{fig:structure_and_stroke}(a) and 514 radicals for 3,755 Level-1 characters. Each character can either be represented as a radical-structured tree or a radical sequence through the depth-first search (see Fig. \ref{fig:structure_and_stroke}(b)). At the stroke level, each Chinese character can be composed by five basic strokes, including \textit{horizontal}, \textit{vertical}, \textit{left-falling}, \textit{right-falling}, and \textit{turning}. There are two or more instances in each category (see Fig. \ref{fig:structure_and_stroke}(c)).

\subsection{Chinese Character Recognition}
According to the different ways of character decomposition, existing CCR methods can be divided into three approaches, \textit{i.e.}, character-based, radical-based, and stroke-based approach.

\noindent\textbf{Character-Based Approach} As convolution neural networks \cite{krizhevsky2012imagenet} thrived over the last decade, CCR has been pushed forward by a large margin. For example, MDCNN \cite{cirecsan2015multi} designs a series of deep neural networks consisting of convolutional blocks, and ensembles their results to produce the final prediction. Zhang et al. \cite{zhang2017online} view traditional features (\textit{e.g.}, Gabor and HoG maps) as priors and incorporated them into neural networks to learn more robust representations. In \cite{xiao2019template}, the authors propose a template-instance loss, which regard printed fonts as prototypes \cite{snell2017prototypical} for constraining features of the same class to be close in the latent space. Generally, the character-based approach is sensitive to categorical distribution changes and usually fails to recognize those characters that have not appeared during training, namely the character zero-shot problem.

\noindent\textbf{Radical-Based Approach} In the past five years, some methods try to represent each Chinese character at the radical level to alleviate the limitations induced by the difference of categorical distribution \cite{li2018building,zhou2016discriminative,tang2017semi,xu2019multiple,cheng2016handwritten,zhong2016handwritten}. For instance, inspired by image captioning, DenseRAN \cite{wang2018denseran} takes the first attempt to iteratively generate the radical sequence of a given character, which are matched with a collection of pre-defined ideographic description sequences (IDS) using edit distance to yield the final character prediction. HDE \cite{cao2020zero} manually designs embeddings for each character using radical-composition knowledge while following the embedding-matching strategy for prediction in the test stage. Albeit these methods can alleviate the problems caused by the difference of categorical distribution, as the distribution discrepancy further increases, the radical zero-shot problem will arise if some uncommon radicals are absent from the training dataset.

\noindent\textbf{Stroke-Based Approach} To fundamentally address the problems incurred by the categorical distribution discrepancy, Chen et al. \cite{chen2021zero} decompose each Chinese character into stroke sequence. However, representing characters at the stroke level will lead to a severe one-to-many problem as each stroke sequence may correspond to many characters. Although it shows superiority in addressing the zero-shot problems, its performance drops obviously as the discrepancy of visual distribution increases.

\section{Methodology}
The architecture of our method is shown in Fig. \ref{fig:architecture_overall}. Specifically, given an image $\mathbf{X} \in \mathbb{R}^{H \times W \times C}$, we utilize an encoder based on ResNet-34 \cite{he2016deep} to extract the feature maps $\mathbf{F} \in \mathbb{R}^{H/2 \times W/2 \times C^{\prime}}$, which are sequentially sent to two decoders and a translator to yield the character-level predictions. Different from the previous methods focusing on a single level, we represent each character as a RSST to fully exploit the merits at different levels.

\subsection{Feature-to-Radical Decoder}The Feature-to-Radical Decoder (FRD) aims to decompose each character at the radical level. Specifically, we build this decoder based on Transformer \cite{vaswani2017attention}, which recently proves to be effective in vision tasks \cite{dosovitskiy2020image}. The feature maps $\mathbf{F} \in \mathbb{R}^{H/2 \times W/2 \times C^{\prime}}$ generated by the encoder are first reshaped to $\tilde{\mathbf{F}} \in \mathbb{R}^{HW/4 \times C^{\prime}}$, then fed to the Transformer along with the right-shifted radical level label following the force teaching manner used in \cite{shi2018aster}. Through $N$ Transformer layers (see detailed configurations in Supplementary Material), the input right-shifted sequence is transformed into the predicted radical sequence. In fact, we can simply calculate the cross-entropy loss to supervise the radical-level predictions. However, due to the severe class imbalanced problem induced by the radical-level decomposition mentioned in \cite{chen2021zero}, we observe that FRD is weak to recognize those low-frequency radicals. In this case, the generated attention maps cannot precisely correspond to the position of the radical at each time step, which brings difficulties to the downstream Radical-to-Stroke Decoder (see more discussions in Supplementary Material). Therefore, we replace the explicit radicals of original radical-level labels with their relative positions in the corresponding radical tree for implicit supervision (see ``Radical GT $\mathbf{\hat{R}}$'' in Fig. \ref{fig:architecture_overall}). We denote the modified radical-level label as $\mathbf{\hat{R}}$ and the predicted sequence as $\mathbf{R}$. Finally, the FRD is supervised by the cross-entropy loss as follows:
\begin{linenomath*}
\begin{equation}
    l_{\text{FRD}}=
    \text{CrossEntropy($\mathbf{R}$,$\mathbf{\hat{R}}$)}
\end{equation}
\end{linenomath*}

\subsection{Radical-to-Stroke Decoder}

The Radical-to-Stroke Decoder (RSD) aims to further decompose FRD’s intermediate outputs (\textit{i.e.}, features of radical regions) into stroke sequences. The RSD is also built upon Transformer, whose architecture is the same as that used in the FRD. Through the FRD, a sequence of attention maps $\mathbf{A} = (\mathbf{a}_{1}, \mathbf{a}_{2}, ..., \mathbf{a}_{T})$ are generated, where $\mathbf{a}_{*} \in \mathbb{R}^{HW/4}$ and $T$ is the maximum length of the sequence. We simply multiply the flattened feature map $\tilde{\mathbf{F}}$ and $\mathbf{A}$ to yield the radical-focused features $\mathbf{F}^{\prime} = (\mathbf{F}^{\prime}_{1}, \mathbf{F}^{\prime}_{2}, ..., \mathbf{F}^{\prime}_{T})$, where $\mathbf{F}^{\prime}_{*} \in \mathbb{R}^{HW/4 \times C^{\prime}}$. To further represent Chinese characters at the stroke level, we feed the right-shifted stroke-level label $\mathbf{\hat{S}}_\text{right}$ and radical-focused features $\mathbf{F}^{\prime}_{*}$ to the RSD for generating the stroke-level prediction $\mathbf{S} = (\mathbf{S}_{1}, \mathbf{S}_{2}, ..., \mathbf{S}_{T})$. We denote the ground truth as $\mathbf{\hat{S}} = (\mathbf{\hat{S}}_{1}, \mathbf{\hat{S}}_{2}, ..., \mathbf{\hat{S}}_{T})$ and define $\mathbf{\hat{S}}_{t}$ at the $t$-th time step  in the following two ways: (1) If it corresponds to a \textit{radical structure}, we concatenate the stroke sequence of all its leaf nodes in the depth-first-search order as $\mathbf{\hat{S}}_{t}$. (2) If it corresponds to a \textit{radical}, we simply set $\mathbf{\hat{S}}_{t}$ to the stroke sequence of the radical (see ``Stroke GT $\hat{\mathbf{S}}$'' in Fig. \ref{fig:architecture_overall}). In this way, the attention maps of radical structures can cover all the regions of its leaf nodes so as to yield more accurate radical-structure predictions. Finally, the RSD is supervised by cross-entropy loss:
\begin{linenomath*}
\begin{equation}
    l_{\text{RSD}}=
    \text{CrossEntropy($\mathbf{S}$,$\mathbf{\hat{S}}$)}
\end{equation}
\end{linenomath*}

Based on these preparations, we integrate the radical structures generated in the FRD and the stroke sequences of the leaf nodes generated in the RSD, to construct a RSST $\mathbf{M}$ for representing Chinese characters (see Fig. \ref{fig:architecture_overall}). The way for tree-to-character translation will be introduced in Sec. \ref{tct}.

\begin{figure}[t]
    \centering
    \includegraphics[width=0.43\textwidth]{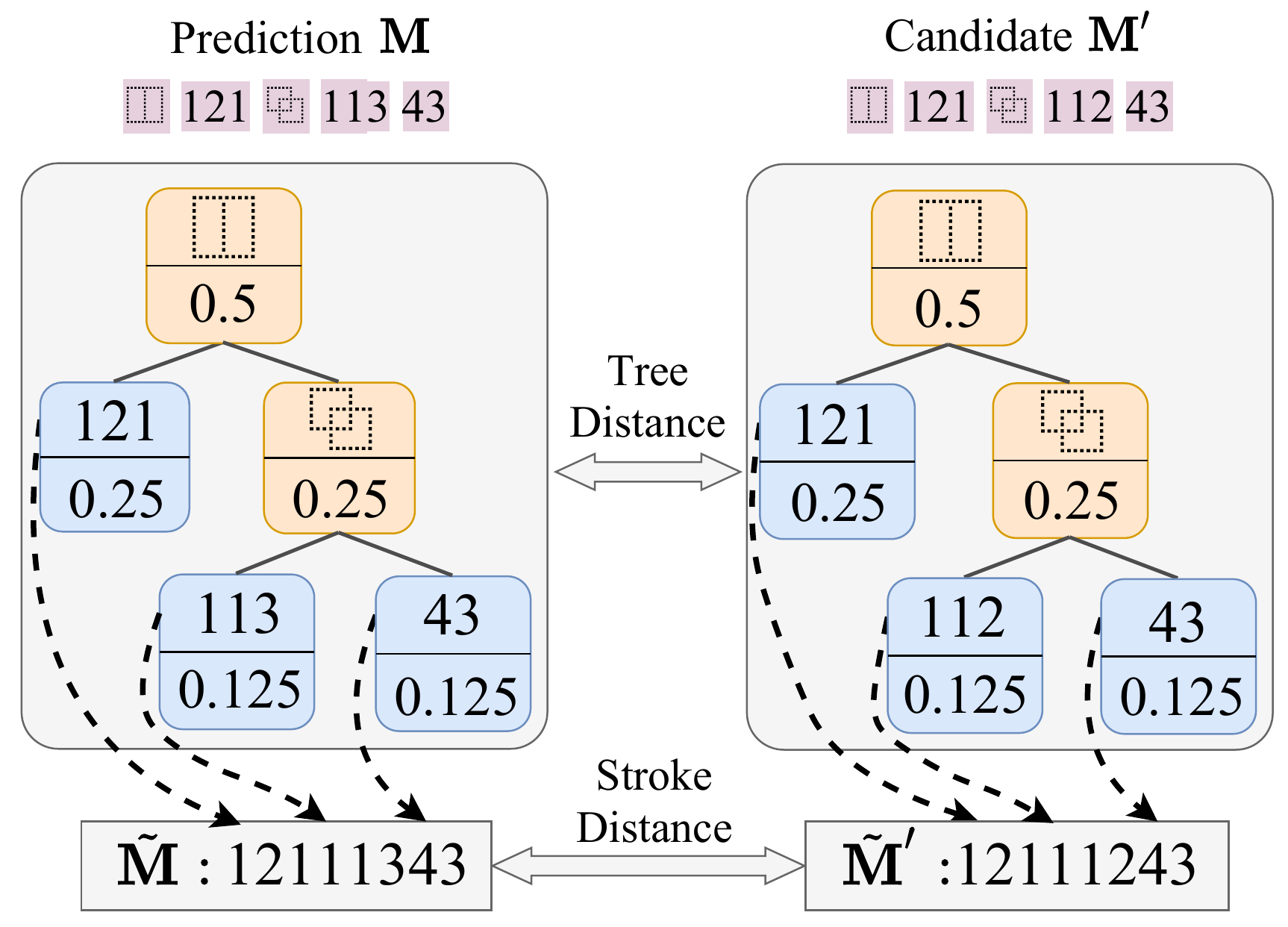}
    \caption{The illustration of tree distance and stroke distance.}
    \label{fig:Metric}
\end{figure}

\subsection{Overall Loss Function}
The overall loss function can be calculated as follows:
\begin{linenomath*}
\begin{equation}
    l = l_{\text{RSD}} + \lambda l_{\text{FRD}}
\end{equation}
\end{linenomath*}
where $\lambda$ is a hyperparameter to balance the two loss functions, which will be discussed in Sec. \ref{sec:choice}.

\subsection{Tree-to-Character Translator}
\label{tct}
We design a Tree-to-Character Translator (TCT), which is only used during testing, to transform the generated tree into a specific character through rectification and matching (\textit{i.e.}, Distance Calculator and Similarity Calculator in Fig. \ref{fig:architecture_overall}). 

\noindent\textbf{Rectification} Since the predicted tree $\mathbf{M}$ may fail to match any trees in the pre-defined RSST lexicon $\mathcal{L}_{\text{tree}}$ that contains tree representation of all characters, it is necessary to rectify $\mathbf{M}$ to match the closest candidate. For the stroke-based method \cite{chen2021zero}, the authors simply used the edit distance to rectify the predicted stroke sequence. In fact, the stroke distance cannot serve as a robust metric since it ignores the spatial cues (\textit{i.e.}, radical structures). An example for illustration is shown in Supplementary Materials.

To mitigate the aforementioned problems, we propose a Weighted Edit Distance (WED) that takes the tree structure of the predicted sequence into account. Inspired by HDE \cite{cao2020zero}, we assign different weights for the radical structures and stroke sequences according to their levels in the tree. Practically, for an element at the $k$-th level, we set its weight to $\alpha ^k$ (see Fig. \ref{fig:Metric}), where $\alpha$ is empirically set to 0.5. To calculate the distance between $\mathbf{M}$ and a candidate $\mathbf{M}^{\prime}$ in the RSST lexicon $\mathcal{L}_{\text{tree}}$, we define the state transition equation $DP$ for dynamic programming as follows:
\begin{linenomath*}
\begin{equation}
DP[i, j]=\text{min} \left\{
\begin{array}{l}
DP[i-1, j]+w_i, \\
DP[i, j-1]+w^{\prime}_j, \\
DP[i-1, j-1]+\frac{ED(\mathbf{M}_i, \mathbf{
M}^{\prime}_j)\times w^{\prime}_j}{Len(\mathbf{M}^{\prime}_j)}
\end{array}
\right.
\label{3}
\end{equation}
\end{linenomath*}
where $DP[i,j]$ denotes the distance between the first $i$ elements of $\mathbf{M}$ and the first $j$ elements of $\mathbf{M}^{\prime}$ in the depth-first-search sequence. $w_i$ and $w^{\prime}_j$ are the corresponding weights for the element $\mathbf{M}_{i}$ and $\mathbf{M}^{\prime}_{j}$, respectively. As the initial state, $DP[0,0]$ is set to 0. $ED$ and $Len$ denote the vanilla edit distance and length of sequence.

Therefore, the WED of the whole tree is calculated as follows (the details and examples are shown in Supplementary Material):
\begin{linenomath*}
\begin{equation}
D_\text{tree}=DP[Len(\mathbf{M}), Len(\mathbf{M}^{\prime})]
\end{equation}
\end{linenomath*}
As shown in Fig. \ref{fig:Metric}, we combine $D_\text{tree}$ and the stroke sequence edit distance $D_\text{stroke}$ used in \cite{chen2021zero} to construct a more comprehensive distance metric $D^{\prime}$:
\begin{linenomath*}
\begin{equation}
D^{\prime} = D_\text{tree} + \beta D_\text{stroke}  \label{4}
\end{equation}
\end{linenomath*}
where $\beta$ is empirically set to 1, and $D_\text{stroke}$ is calculated as follows:
\begin{linenomath*}
\begin{equation}
D_\text{stroke}=ED(Concat(\tilde{\mathbf{M}}), Concat(\tilde{\mathbf{M}}^{\prime}))
\label{6}
\end{equation}
\end{linenomath*}
where $Concat(\cdot)$ represents the concatenation while $\tilde{\mathbf{M}}$ and $\tilde{\mathbf{M}}^{\prime}$ represent the sub-sequences of $\mathbf{M}$ and $\mathbf{M}^{\prime}$ that only contain strokes (see ``Stroke Distance'' in Fig. \ref{fig:Metric}). Finally, the tree candidate having the minimum $D^{\prime}$ is chosen as the rectified prediction $\mathbf{M}_\text{rec}$. Please note that if the predicted $\mathbf{M}$ can precisely match a candidate in $\mathcal{L}_{\text{tree}}$, we directly set $\mathbf{M}_\text{rec} = \mathbf{M}$.

\noindent\textbf{Matching} Since the rectified tree $\mathbf{M}_{\text{rec}}$ may correspond to many characters (\textit{i.e.}, the one-to-many problem), we also utilize a Siamese architecture, as designed in \cite{chen2021zero}, to match the rectified prediction $\mathbf{M}_\text{rec}$ with a specific character to further improve the performance. We collect a confusable set $\mathcal{C}$ containing those tree sequences that can match more than one character (see Fig. \ref{fig:architecture_overall}). Compared with \cite{chen2021zero}, our method contains a smaller confusable set $\mathcal{C}$ since we take both radical structures and strokes into consideration for representing Chinese characters (details are shown in Supplementary Materials). Concretely, among 3,755 Level-1 characters, there are 280 confusable characters for the stroke-based approach and only 111 confusable characters for our radical-structured tree representation.

In the test phase, if $\mathbf{M}_\text{rec}$ can match a specific character ($\mathbf{M}_\text{rec} \notin \mathcal{C}$), the character is directly viewed as the final prediction. Otherwise, we compare the feature maps $\mathbf{F}$ with the features of support samples $\mathcal{S}$ having the same tree representation. Specifically, $\mathcal{S}$ is fed into the same encoder to yield a list of feature maps $\mathcal{F}$, where $N$ is the number of candidates. Then we calculate the \textit{Cosine distance} between $\mathbf{F}$ and the feature maps of each candidate, then choose the candidate with the maximum feature similarity as the final prediction.


\begin{figure}[t]
    \centering
    \includegraphics[width=0.44\textwidth]{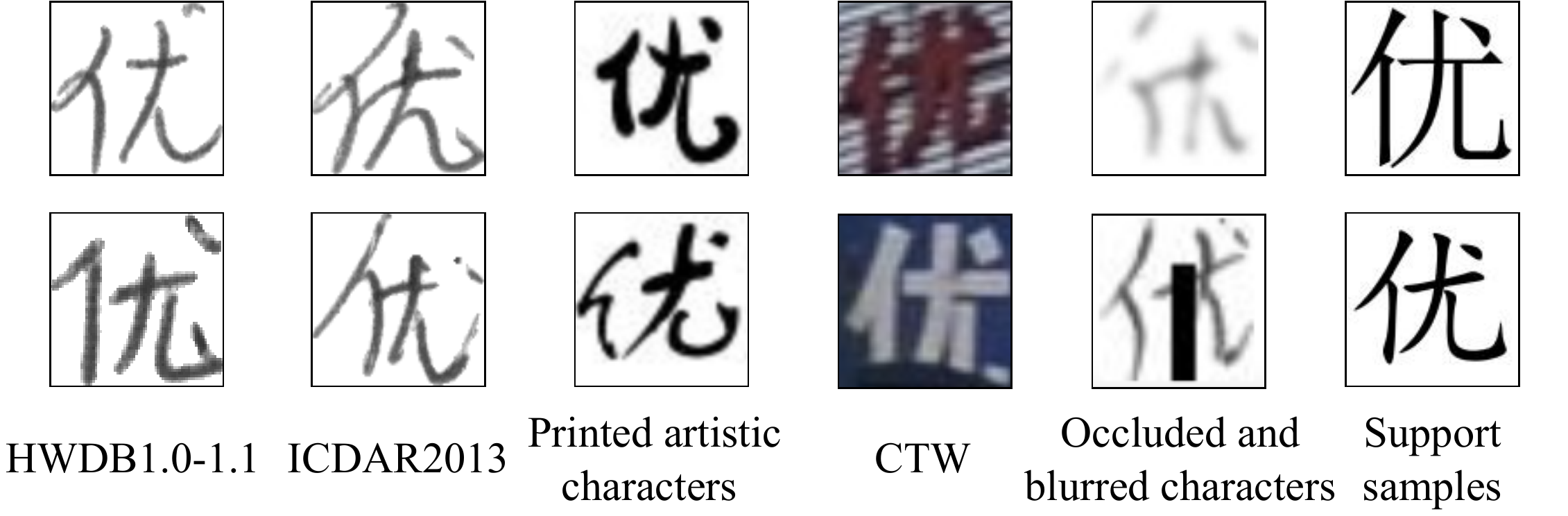}
    \caption{Examples of the character ``You'' in each dataset.}
    \label{fig:examples}
\end{figure}

\section{Experiments}

\subsection{Datasets and Experimental Settings} 
\textbf{Datasets} We conduct the experiments with the following datasets (examples of each dataset are shown in Fig. \ref{fig:examples}).
\begin{itemize}
\item \textbf{HWDB1.0-1.1} \cite{liu2013online} comprises 2,678,424 handwritten Chinese character images written by 720 writers. It contains 3,881 character classes covering all the 3,755 Level-1 characters.
    
\item \textbf{ICDAR2013} \cite{yin2013icdar} is collected from 60 writers (different from the writers of HWDB1.0-1.1 \cite{liu2013online}), containing 224,419 offline handwritten images with 3,755 classes. 
    
\item \textbf{Printed Artistic Characters} \cite{chen2021zero} collects 105 printed artistic fonts to yield 394,275 images for 3,755 classes.
    
\item \textbf{CTW} \cite{yuan2019large} is collected from the street view. It has 760,107 samples for training and 52,765 samples for testing. These samples contain 3,650 classes. 
    
\item \textbf{Occluded and Blurred Characters} are constructed by exerting three degrees (Easy, Medium, and Hard) of occlusion and blur on ICDAR2013 \cite{yin2013icdar}. The details are shown in Supplementary Material.
    
\item \textbf{Support Samples} are generated by two widely-used fonts including Simsun and Simfang (not in the above-mentioned artistic fonts).
\end{itemize}

\noindent\textbf{Details of Zero-Shot Settings}
We follow \cite{chen2021zero} to construct datasets for the zero-shot settings. The configuration of each notation will be detailed in Sec. \ref{sec:choice} and Sec. \ref{sec:exp}.

\textit{Character Zero-Shot Setting} consists of several sub-settings controlled by the parameter $m$. Concretely, we collect samples of the first \textit{m} classes in the alphabet $\mathcal{A}$ from dataset $\mathcal{D}_{\text{A}}$ for training while collecting samples of the last $N_{\text{last}}$ classes in dataset $\mathcal{D}_{\text{B}}$ for testing.

\textit{Radical Zero-Shot Setting} consists of several sub-settings controlled by the parameter $n$. According to the frequency of each radical in 3,755 Level-1 characters, if a character with a radical that appears more than \textit{n} times, the samples of this character in $\mathcal{D}_{\text{A}}$ are used for training, otherwise those samples in $\mathcal{D}_{\text{B}}$ are used for testing.

\begin{table}[t]
\renewcommand{\arraystretch}{1.1}
\caption{Choices of $\lambda$ to balance $l_{\text{FRD}}$ and $l_{\text{RSD}}$.}
\centering
\scalebox{0.90}{
\begin{tabular}{cccccc}
\hline
\multirow{2}*{$\lambda$}  &  \multicolumn{5}{c}{$m$ for Handwritten Character Zero-Shot} \\
\cline{2-6} 
~ & 500 & 1000 & 1500 & 2000 & 2755 \\ 
\hline
0.01 & 5.67\% & 21.62\% & 32.99\% & 37.60\% & 33.93\% \\
0.1 & \textbf{11.56}\% & 21.83\% & \textbf{35.32}\% & \textbf{39.22}\% & \textbf{47.44}\% \\
1 & 10.99\% & \textbf{23.92}\% &  30.97\% &  36.96\% &  42.56\%    \\
10 &  4.60\% &   12.63\% &  19.66\% & 24.16\% &  33.11\% \\
\hline
\end{tabular}}
\label{tab:parameter sensibility}
\end{table}

\begin{table}[t]
\renewcommand{\arraystretch}{1.1}
\caption{Choices of $D_{\text{tree}}$ and $D_{\text{stroke}}$ for the distance metric.}
\centering
\scalebox{0.9}{
\begin{tabular}{cccccc}
\hline
\multirow{2}*{\textbf{Method}}  &  \multicolumn{5}{c}{$m$ for Handwritten Character Zero-Shot} \\
\cline{2-6} 
~ & 500 & 1000 & 1500 & 2000 & 2755 \\ 
\hline
w/o $D_{\text{tree}}$ & 8.51\% & 17.53\% & 29.95\% & 33.52\% & 41.51\% \\
w/o $D_{\text{stroke}}$ & 10.90\% & 18.86\% & 28.20\% & 32.20\% & 39.92\% \\
Ours & \textbf{11.56}\% & \textbf{21.83}\% & \textbf{35.32}\% & \textbf{39.22}\% & \textbf{47.44}\% \\
\hline
\end{tabular}}
\label{tab:ablation}
\end{table}

\begin{table*}[t]
\renewcommand{\arraystretch}{1.1}
\caption{The experimental results of character zero-shot (left) and radical zero-shot (right) settings.
}
\centering
\scalebox{0.80}{
\begin{tabular}{cccccc|ccccc}
\hline
\multirow{2}*{\textbf{Handwritten}}  & \multicolumn{5}{c|}{ $m$ for Character Zero-Shot Setting}  & \multicolumn{5}{c}{ $n$ for Radical Zero-Shot Setting} \\
\cline{2-11} 
~ & 500 & 1000  & 1500 & 2000 & 2755 & 50 & 40  & 30 & 20 & 10 \\
\hline
DenseRAN \cite{wang2018denseran} & 1.70\% & 8.44\% & 14.71\% & 19.51\% & 30.68\% & 0.21\% & 0.29\% & 0.25\% & 0.42\% & 0.69\% \\
HDE \cite{cao2020zero} & 4.90\% & 12.77\% & 19.25\% & 25.13\% & 33.49\% & 3.26\% & 4.29\% & 6.33\% & 7.64\% & 9.33\%\\
Chen et al. \cite{chen2021zero} & 5.60\% & 13.85\% & 22.88\% & 25.73\% & 37.91\% & 5.28\% & 6.87\% & 9.02\% & 14.67\% & 15.83\%\\
Ours & \textbf{11.56\%} & \textbf{21.83\%} & \textbf{35.32\%} & \textbf{39.22\%} & \textbf{47.44\%} & \textbf{7.94\%} & \textbf{11.56\%} & \textbf{15.13\%} & \textbf{15.92\%} & \textbf{20.21\%}\\
\hline
\hline
\multirow{2}*{\textbf{Printed Artistic}}  & \multicolumn{5}{c|}{ $m$ for Character Zero-Shot Setting} & \multicolumn{5}{c}{
 $n$ for Radical Zero-Shot Setting} \\
\cline{2-11} 
~ & 500 & 1000  & 1500 & 2000 & 2755 & 50 & 40  & 30 & 20 & 10 \\ \hline
DenseRAN \cite{wang2018denseran} & 0.20\% & 2.26\% & 7.89\% & 10.86\% & 24.80\% & 0.07\% & 0.16\% & 0.25\% & 0.78\% & 1.15\% \\
HDE \cite{cao2020zero} & 7.48\% & 21.13\% & 31.75\% & 40.43\% & 51.41\% & 4.85\% & 6.27\% & 10.02\% & 12.75\% & 15.25\%\\
Chen et al. \cite{chen2021zero} & 7.03\% & 26.22\% & 48.42\% & 54.86\% & 65.44\% & 11.66\% & 17.23\% & 20.62\% & 31.10\% & 35.81\%\\
Ours & \textbf{23.12}\% & \textbf{42.21\%} & \textbf{62.29\%} & \textbf{66.86\%} & \textbf{71.32\%} & \textbf{13.90\%} & \textbf{19.45\%} & \textbf{26.59\%} & \textbf{34.11\%} & \textbf{38.15\%}\\
\hline
\hline
\multirow{2}*{\textbf{Scene}}  & \multicolumn{5}{c|}{ $m$ for Character Zero-Shot Setting} & \multicolumn{5}{c}{
 $n$ for Radical Zero-Shot Setting} \\
\cline{2-11}
~ & 500 & 1000  & 1500 & 2000 & 3150 & 50 & 40  & 30 & 20 & 10 \\ \hline
DenseRAN \cite{wang2018denseran} & 0.15\% & 0.54\% & 1.60\% & 1.95\% & 5.39\% & 0\% & 0\% & 0\% & 0\% & 0.04\% \\
HDE \cite{cao2020zero} & 0.82\% & 2.11\% & 3.11\% & 6.96\% & 7.75\% & 0.18\% & 0.27\% & 0.61\% & 0.63\% & 0.90\%\\
Chen et al. \cite{chen2021zero} & \textbf{1.54}\% & \textbf{2.54}\% & 4.32\% & 6.82\% & 8.61\% & 0.66\% & 0.75\% & 0.81\% & 0.94\% & 2.25\% \\
Ours & 1.41\% & 2.53\% & \textbf{4.59\%} & \textbf{9.32\%} & \textbf{13.02\%} & \textbf{1.21\%} & \textbf{1.29\%} & \textbf{1.89\%} & \textbf{2.90\%} & \textbf{3.88\%} \\
\hline
\end{tabular}}

\label{tab:big table}
\end{table*}

\subsection{Implementation Details} 
We implement our method with PyTorch. The training and evaluation of all subsequent experiments are both conducted on an NVIDIA RTX 2080Ti GPU with 11GB memory. The Adadelta optimizer \cite{zeiler2012adadelta} is utilized with the learning rate set to 0.1. The batch size is set to 8. Each input image is resized to 32 × 32 and normalized to [-1,1]. We only use one Transformer layer ($N=1$) in the FRD and RSD. Please note that no model ensemble and data augmentation strategies are used. We train the model with the proposed RSST representation for all settings. Then, in the standard settings (\textit{i.e.}, all settings except the zero-shot settings), the model is fine-tuned with character-level labels.

\subsection{Choices of Parameters} \label{sec:choice}
We conduct the following experiments using HWDB1.0-1.1 \cite{liu2013online} in the character zero-shot setting. Specifically, we set $\mathcal{D}_{A}$ and $\mathcal{D}_{B}$ to HWDB1.0-1.1 and ICDAR2013, respectively. The alphabet $\mathcal{A}$ is set to 3,755 Level-1 characters and $N_{\text{last}}$ is set to 1,000. Please note that all experiments in Sec. \ref{sec:exp} are based on the same selected hyperparameters.

\noindent\textbf{Choices of $\lambda$} To  deeply dive into the weight $\lambda$ for the overall loss $l$, we explore $\lambda$ from \{0.01, 0.1, 1, 10\}. As shown in Tab. \ref{tab:parameter sensibility}, when $\lambda=0.1$, the performance surpasses all its counterparts in all cases. Interestingly, we observe that radical structures are easier to distinguish than strokes as the radical structures are visually apparent. Hence, we give the radical-structure loss $l_{\text{FRD}}$ a lower weight (set $\lambda$ to 0.1) in the following experiments.

\noindent\textbf{Choices of $D_{\text{tree}}$ and $D_{\text{stroke}}$} We try to investigate the effectiveness of $D_{\text{tree}}$ and $D_{\text{stroke}}$ for the overall distance metric $D^{\prime}$. As shown in Tab. \ref{tab:ablation}, both of the two terms perform crucial roles for the matching process. Specifically, our method outperforms previous SOTA methods by almost 10\% when $m = 2,755$. When $D_{\text{tree}}$ or $D_{\text{stroke}}$ is removed from the distance metric, the performance is declined by a large margin. Even so, the declined performance is still better than previous methods focusing on a single-level representation.

\subsection{Experimental Results} \label{sec:exp}
We conduct experiments on the abovementioned datasets. Visualizations of experimental results are shown in Supplementary Materials.

\noindent\textbf{Experiments on Handwritten Characters}  We set $\mathcal{D}_{A}$ to the HWDB1.0-1.1 dataset and $\mathcal{D}_{B}$ to ICDAR2013. We set the alphabet $\mathcal{A}$ to 3,755 Level-1 characters and $N_{\text{last}}$ to 1,000. As shown in the top-left of Tab. \ref{tab:big table}, the proposed method outperforms the previous methods by a large margin in the character zero-shot and radical zero-shot settings. More specifically, when $m$ is set to 2,755, the performance of our method is 9.53\% better than the stroke-based approach \cite{chen2021zero} and 16.76\% better than the radical-based approach \cite{wang2018denseran}. Although the various handwritten styles may result in visual distribution changes, our method can robustly overcome this visual discrepancy even if the characters or radicals are absent from the training dataset. 

\begin{table}[t]
\caption{The experimental results of occluded and blurred characters images that are collected from the CTW dataset.}
\centering
\scalebox{0.80}{
\begin{tabular}{c|p{2.7cm}<{\centering} p{2.5cm}<{\centering}}
\hline
\multirow{2}*{Method}  &  Occluded  & Blurred \\
~ & Characters & Characters\\
\hline
CharRep & 33\% & 50\% \\
CharRep + DataAug & 38\% & 69\% \\
Ours + CharRep & \textbf{44}\% & \textbf{74}\% \\
\hline
\end{tabular}}

\label{tab:Result in CTW}
\end{table}

\begin{figure}[t]
    \centering
    \includegraphics[width=0.42\textwidth]{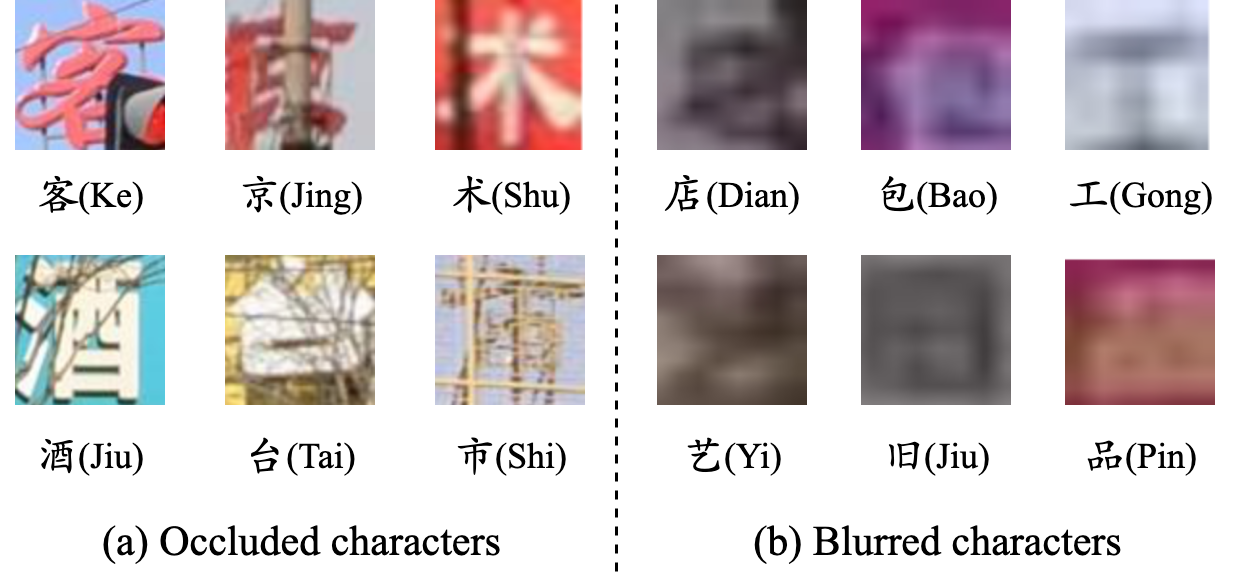}
    \caption{Examples of the occluded and blurred character images selected from the CTW dataset.}
    \label{fig:selected_ctw}
\end{figure}

\noindent\textbf{Experiments on Printed Artistic Characters} We set both $\mathcal{D}_{A}$ and $\mathcal{D}_{B}$ to the whole dataset for the zero-shot settings. We set the alphabet $\mathcal{A}$ to 3,755 Level-1 characters and $N_{\text{last}}$ to 1,000. The image quality of this dataset is relatively clearer (\textit{e.g.}, the strokes are usually easy to distinguish) compared with other datasets. Through the experimental results, the proposed method outperforms previous methods, by a larger margin ranging from 5.88\% to 15.99\% in the character zero-shot setting and from 2.22\% to 5.97\% in the radical zero-shot setting (see the middle row of Tab. \ref{tab:big table}), which further validates the superiority of our RSST representation. 

\begin{table*}[t]
\renewcommand{\arraystretch}{1.1}
\caption{The experimental results of occluded and blurred character settings. ``None'' denotes no blur or occlusion are used for the testing dataset. ``CharRep'' denotes that the model is trained or fine-tuned with character-level representation. More experimental results of other character-based methods are shown in Supplementary Material. 
    }
\centering
\scalebox{0.85}{
\begin{tabular}{c|ccc|ccc|c}
\hline
\multirow{2}*{\textbf{Method}}  & \multicolumn{3}{c|}{Occluded Characters} & \multicolumn{3}{c|}{Blurred Characters} & \multirow{2}*{None}\\
\cline{2-7} 
~ & Hard & Medium & Easy & Hard & Medium & Easy & \\
\hline
DenseRAN \cite{wang2018denseran} & 30.77\% & 39.16\% & 53.17\% & 20.94\% & 44.70\% & 58.15\% & 96.66\% \\
HDE \cite{cao2020zero} & 18.25\% & 24.98\% & 37.69\% & 20.91\% & 36.64\% & 54.88\% & 97.14\%\\
Chen et al. \cite{chen2021zero} & 29.63\% & 35.99\% & 57.62\% & 20.17\% & 46.28\% & 59.22\% & 96.74\%\\
\hline
CharRep & 28.63\% & 37.93\% & 55.37\% & 21.18\% & 45.11\% & 59.28\% & 96.17\%\\
Ours & 29.84\% & 38.86\% & 60.12\% & 23.91\% & 47.36\% & 59.43\% & 96.05\%\\
Ours + CharRep & \textbf{41.59\%} & \textbf{52.84\%} & \textbf{70.45\%} & \textbf{30.17\%} & \textbf{52.70\%} & \textbf{64.07\%} & \textbf{97.42\%}\\
\hline
\end{tabular}}

\label{tab:occluded and blurred}
\end{table*}

\noindent\textbf{Experiments on Scene Characters} We set both $\mathcal{D}_{A}$ and $\mathcal{D}_{B}$ to the whole dataset for the zero-shot settings. We sort the 3,650 characters contained in CTW according to their positions in the national standard GB18030-2005 to construct the alphabet $\mathcal{A}$ and set $N_{\text{last}}$ to 500. Different from the handwritten and printed artistic datasets, most samples in CTW have low resolution and complicated backgrounds. As shown in the bottom of Tab. \ref{tab:big table}, the proposed method outperforms the previous methods in most cases. Generally, the performance on scene characters still has room for improvement since scene character images are always accompanied by complex backgrounds, blurring, and occlusions.

\noindent\textbf{Experiments on Occluded and Blurred Characters} We set $\mathcal{D}_A$ to HWDB1.0-1.1 and $\mathcal{D}_B$ to ICDAR2013 with different degrees of blurs and occlusions. As shown in Tab. \ref{tab:occluded and blurred}, we observe that our model outperforms other methods, which validates the strong ability of the proposed RSST representation. Moreover, after the model is pre-trained with the RSST representation, we attempt to fine-tune the model with character-level labels to further take advantage of the merits of character level. Specifically, we simply enlarge the output size of the FRD and set the maximum output length to 1 for directly generating the character-level predictions. We observe that the fine-tuned model is more robust with regard to the challenging scenarios. Moreover, the performance is much better than the character-based approach (refer to ``CharRep'' in Tab. \ref{tab:occluded and blurred}) benefiting from the multi-level knowledge. To evaluate the performance on occluded and blurred characters with real-world deteriorations, we manually collect 100 blurred and 100 occluded samples from the CTW dataset (examples are shown in Fig. \ref{fig:selected_ctw}), which are derived from the real-world scenes. As shown in Tab. \ref{tab:Result in CTW}, after the model is pre-trained with the proposed RSST representation, the performance boosts from 33\% to 44\% (11\% $\uparrow$) on the blurred samples and from 50\% to 74\% (24\% $\uparrow$) on the occluded samples. In contrast, when using data augmentation, the performance improvement is only 5\% on the blurred samples and 19\% on the occluded samples.

\section{Discussions}

\begin{figure}[t]
    \centering
    \includegraphics[width=0.40\textwidth]{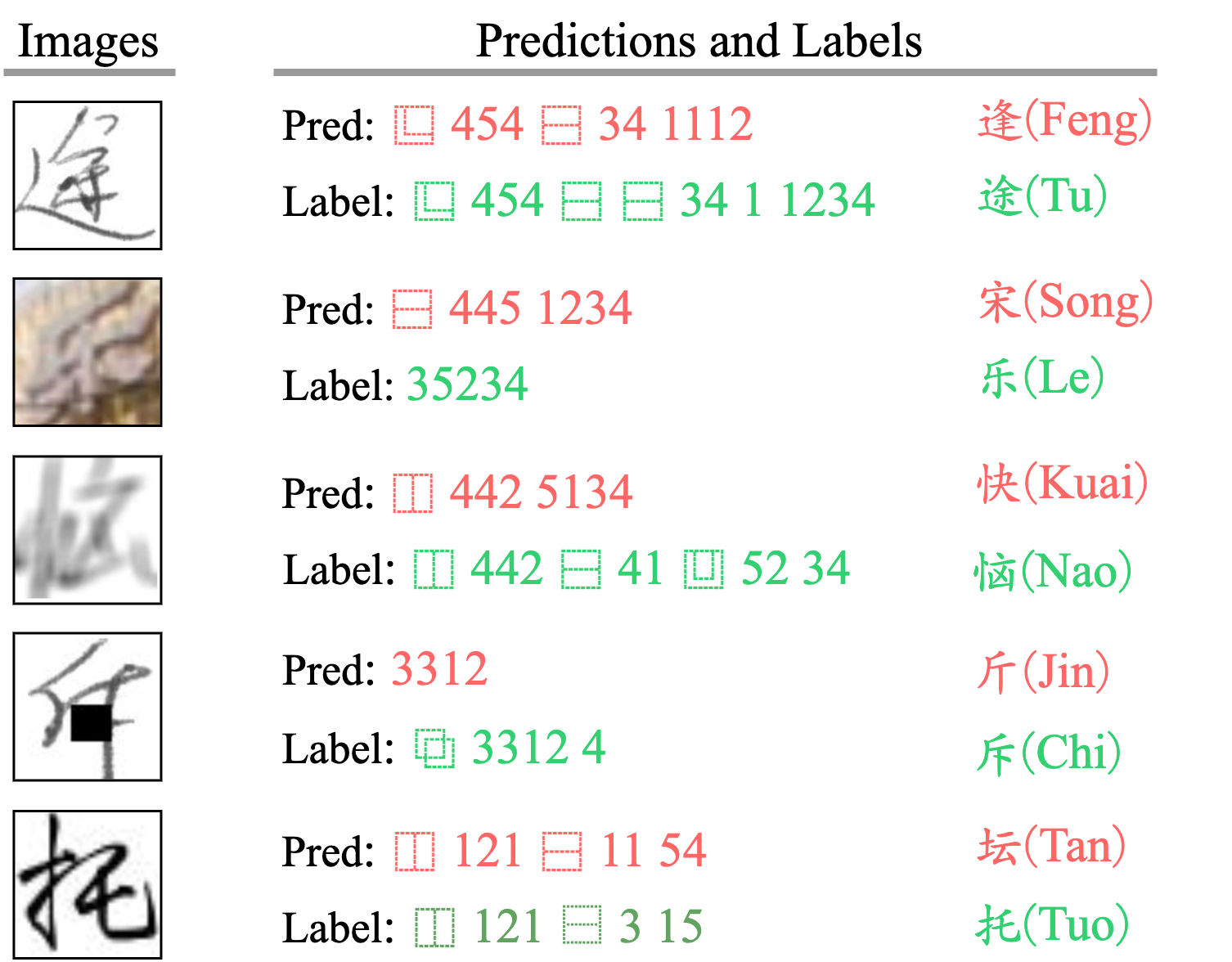}
    \caption{Visualizations of some failure cases.}
    \label{fig:failure cases}
\end{figure}

\noindent\textbf{Failure Cases} Some failure cases are shown in Fig. \ref{fig:failure cases}. We observe that the connected strokes bring difficulties to our method. For the handwritten character ``Tu’’, the left-falling and right-falling strokes at the end of the sequence are merged together, which confuses our model thus yielding the wrong predictions. This situation also exists in the blurred character ``Nao'' and the artistic character ``Tuo''. In particular, the connected strokes are more difficult to identify in a blurred situation. The oblique characters will also hamper the performance of our model (refer to the scene character ``Le''). It is reasonable since the proposed RSD concentrates on the stroke level, thus is sensitive to the rotation angle of the character. For the occluded character ``Chi'', our method mistakenly recognizes this character as ``Jin'' since the right-falling at the end of the sequence is occluded.

\noindent\textbf{Time Efficiency} For fair comparison, we set the batch size to 32 and use ResNet-34 as the encoder and Transformer as the decoder to compare the time efficiency. Specifically, we take the average time for 100 iterations in the test stage. Through the experiments, the methods focusing on a single level cost less time (\textit{e.g.}, 0.06s, 0.33s, 0.65s for the character-based, the radical-based, and the stroke-based methods, respectively) than our method (1.25s). Since our method tries to hierarchically decompose characters at the radical and stroke level to generate more robust tree representations, we sacrifice part of the time efficiency in pursuit of better recognition performance. Furthermore, the time efficiency can be further optimized since the features of all characters in the confusable set $\mathcal{C}$ can be obtained in advance. It is worth mentioning that in the standard settings we only use the proposed RSST representation to pretrain the model and then fine-tune the model at the character level, \textit{i.e.}, the time cost of our method in the standard settings is as the same as the character-based methods.

\section{Conclusions}
In this paper, we propose a new representation called Radical-Structured Stroke Trees (RSST) to tackle CCR. A decomposition framework is put forward to generate the radical structures and strokes of each radical, which can be further organized into a RSST. Furthermore, we propose a Tree-to-Character Translator with the designed Weighted Edit Distance to match a specific tree in the RSST lexicon. The experimental results validate that our method outperforms the SOTA single-level methods by increasing margins as the distribution difference becomes more severe in the blurring, occlusion, and zero-shot scenarios, which validates the superiority of the proposed RSST representation.

\newpage
\bibliography{aaai23}
\newpage
\appendix

\onecolumn
\section{Configurations for Transformer Layer}
\begin{figure*}[h]
    \centering
    \includegraphics[width=0.8\textwidth]{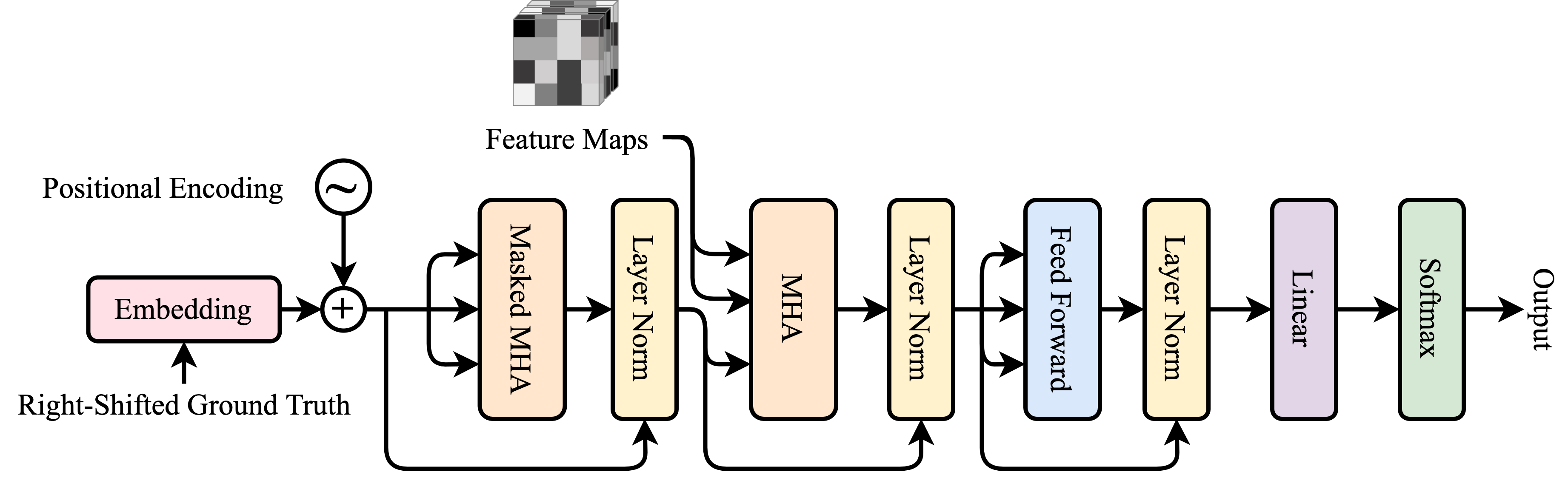}
    \caption{The architecture of the Transformer layer. It follows the basic design of the original Transformer decoder.}
    \label{fig:transformer}
\end{figure*}

In this section, we introduce the configurations of the Transformer layer used in both the Feature-to-Radical Decoder and Radical-to-Stroke Decoder. The Transformer layer consists of three modules, including the masked multi-head attention module (Masked MHA), the multi-head attention module (MHA), and the feed-forward module, the architecture of which is shown in Fig. \ref{fig:transformer}. Following \cite{yang2020holistic}, we only use one Transformer layer for each decoder, and some hyperparameters of this layer are shown in Tab. \ref{transformer_config}.

\begin{table}[h]
\centering
\caption{Hyperparameters and the corresponding values in the Transformer layer.
}
\scalebox{1.0}{
\begin{tabular}{cc}
\hline
Hypermeter & Value \\ 
\hline
The number of decoder blocks & 1 \\
Number of heads & 4 \\
Dimensionality of positional encoding & 1024 \\ 
Dimensionality of embedding & 1024 \\
\hline
\end{tabular}}
\label{transformer_config}
\end{table}

\section{Comparison between Edit Distance and Weighted Edit Distance}
\begin{figure}[h]
    \centering
    \includegraphics[width=0.6\textwidth]{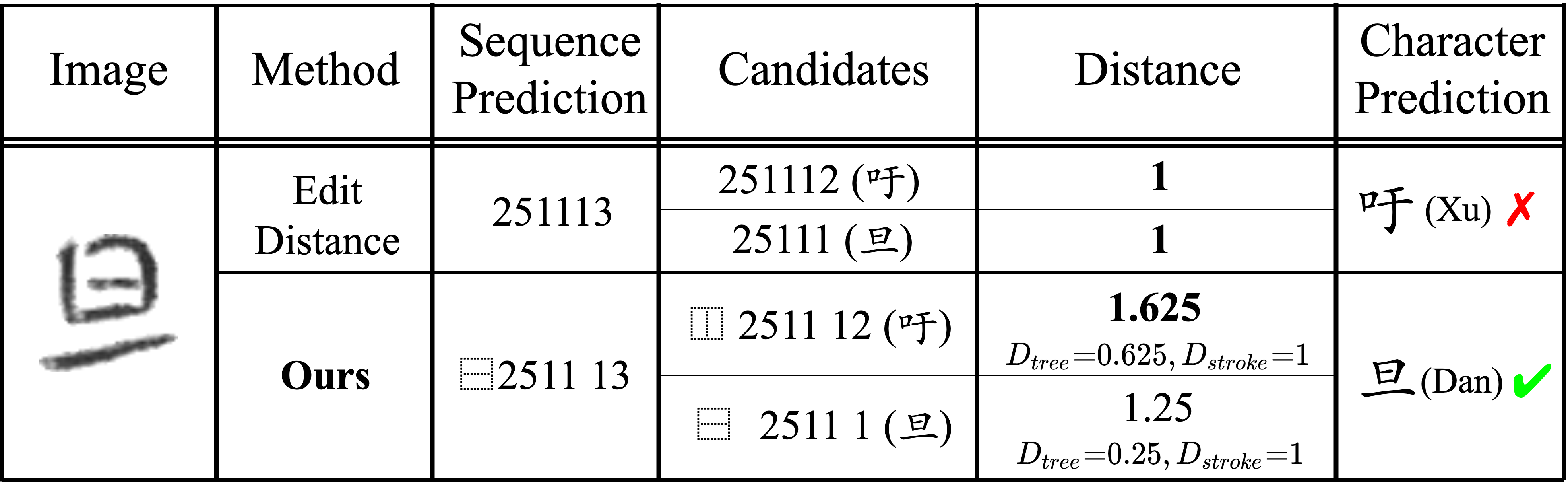}
    \caption{Our metric can robustly rectify the predicted sequence as it takes both the radical structures and strokes into account.}
    \label{fig:method_comparision}
\end{figure}

As shown in Fig. \ref{fig:method_comparision}, Chen et al. \cite{chen2021zero} simply chose the first candidate ``Xu'' as the final prediction, where the edit distance is employed as the metric. When the radical structures are taken into account in the proposed weighted edit distance, the character ``Dan'' is more reasonable.

\section{Supervision for Feature-to-Radical Decoder}
\begin{figure}[h]
    \centering
    \includegraphics[width=0.9\textwidth]{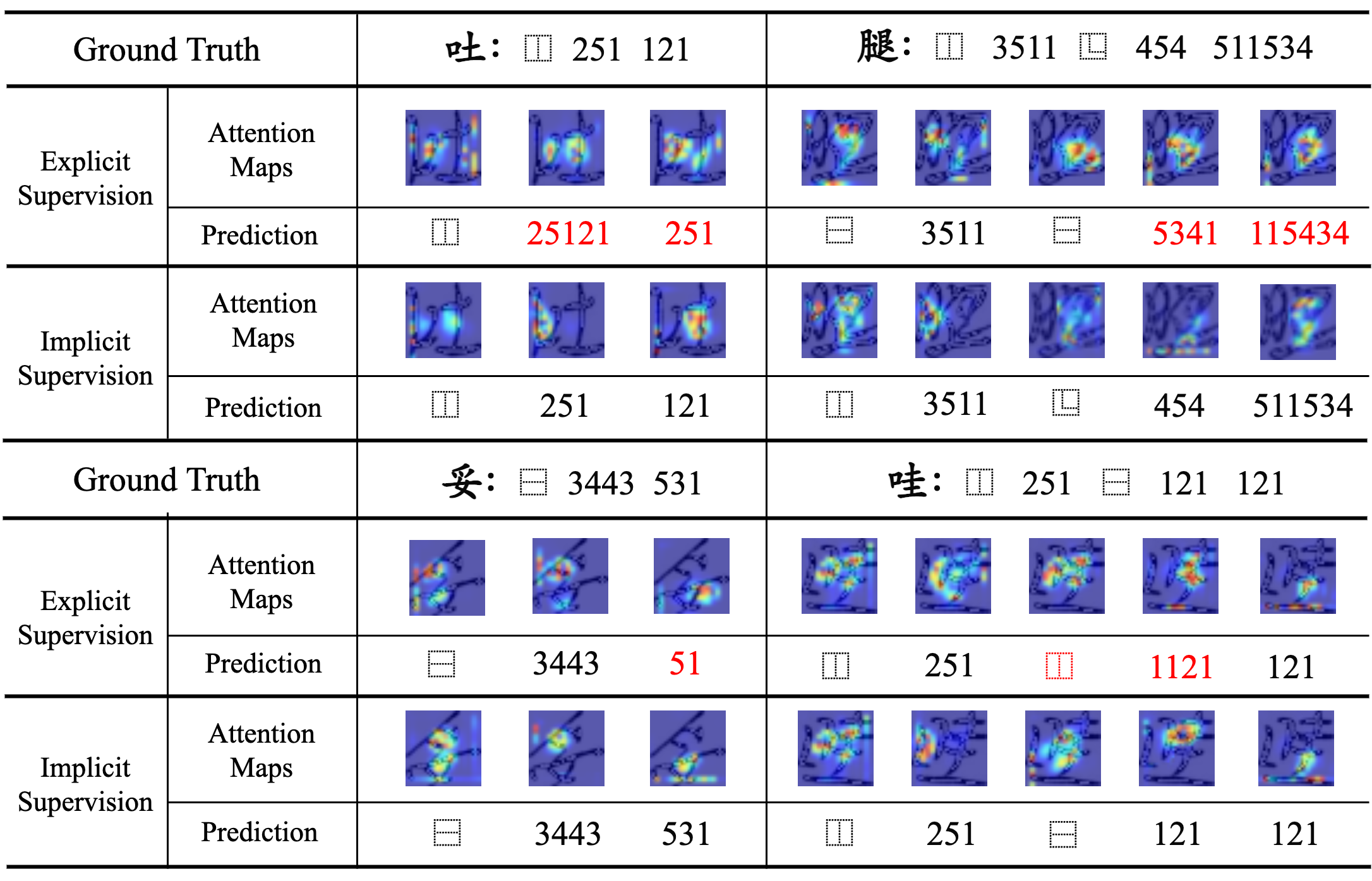}
    \caption{Comparison between implicit supervision and explicit supervision. When using the implicit orders for supervision, the attention maps can better correspond to the position of radical-structure or radical at each time step.}
    \label{fig:att_cmp}
\end{figure}

The Feature-to-Radical Decoder (FRD) aims to generate the radical-structures and the position of each radical (\textit{i.e.}, the generated attention maps). Due to the severe class imbalance problem induced by the radical-level decomposition, the FRD will be weak to recognize those low-frequency radicals and fail to perceive their positions if we utilize the explicit radicals for supervision. Therefore, we replace the explicit radicals with their orders of appearance in the radical sequence for implicit supervision. We conduct experiments to explore the impact of explicit radical supervision and implicit order supervision. As shown in Tab. \ref{tab:radical-supervise}, when using explicit radicals for supervision, the performance declines by a large margin, which stems from the drifted attention maps of radicals. (see Fig. \ref{fig:att_cmp}).

\begin{table}[h]
\centering
\caption{Performance comparison between explicit and implicit supervision.}
\scalebox{0.96}{
\begin{tabular}{crrrrr}
\hline
\multirow{2}*{Supervision}  &  \multicolumn{5}{c}{$m$ for Handwritten Character Zero-Shot} \\
\cline{2-6} 
~ & 500 & 1000 & 1500 & 2000 & 2755 \\ 
\hline
Explicit Supervision & 5.43\% & 12.35\% & 19.56\% & 27.43\% & 32.00\% \\
Implicit Supervision & \textbf{11.74}\% & \textbf{22.91}\% & \textbf{36.33}\% & \textbf{40.27}\% & \textbf{48.00}\%  \\
\hline
\end{tabular}}

\label{tab:radical-supervise}
\end{table}

\section{Preparation for Occluded and Blurred Character Datasets}
\begin{figure}[h]
    \centering
    \includegraphics[width=0.5\textwidth]{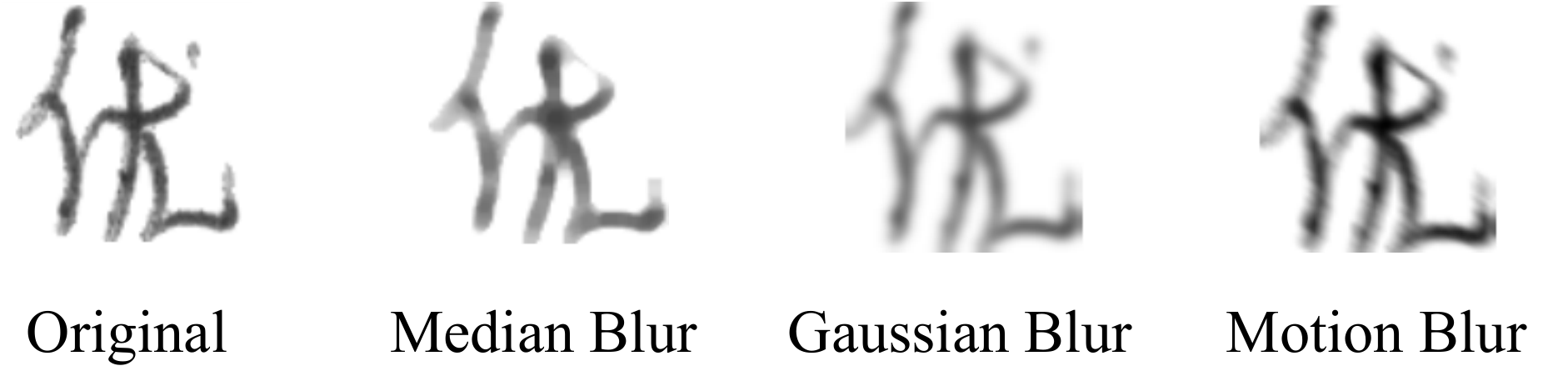}
    \caption{The examples of the three blur operations, including Median Blur, Gaussian Blur, and Motion Blur.}
    \label{fig:blur_example}
\end{figure}

\begin{figure}[h]
    \centering
    \includegraphics[width=0.8\textwidth]{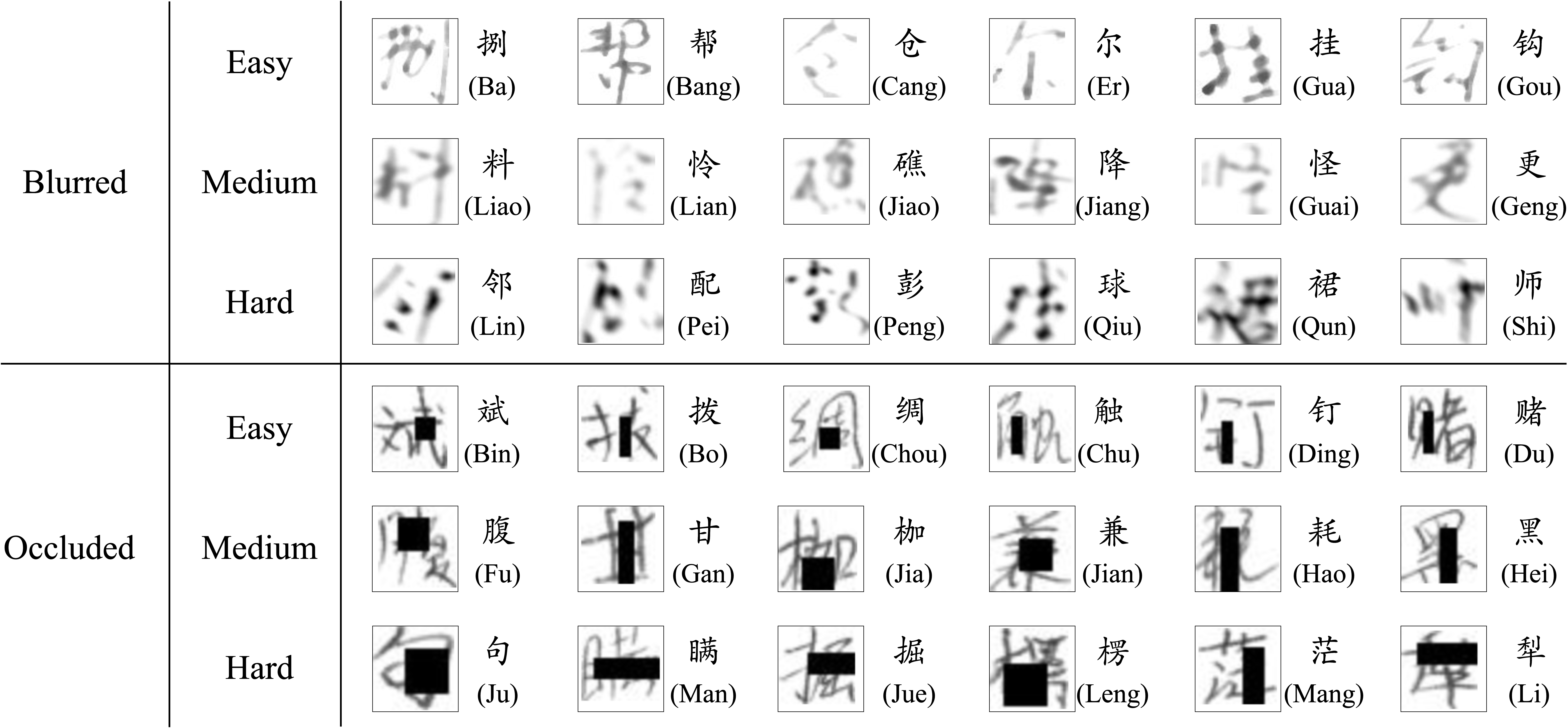}
    \caption{Some examples of occluded and blurred datasets. ``Easy'', ``Medium'', and ``Hard'' represent the different degrees of the occluded or blurred noise.}
    \label{fig:occluded_blurred}
\end{figure}

In this section, we introduce the way of constructing the occluded and blurred character datasets based on ICDAR2013 \cite{yin2013icdar}. For occluded character datasets, we exert different degrees (Easy, Medium, and Hard) of occlusion on the samples to generate the corresponding dataset. Specifically, we utilize a rectangular block of area $n$ to mask the character image, where $n$ is the number of pixels and ranges in \{64, 144, 256\} representing Easy, Medium, and Hard respectively. For blurred character datasets, we utilize three types of blurred operations, including Median Blur, Gaussian Blur, and Motion Blur (the example of each blur operation is shown in Fig. \ref{fig:blur_example}). Concretely, we generate Easy blurred dataset by exerting Median Blur, Medium by exerting Median Blur and Gaussian Blur, and Hard by all of the three blurred operations. Some examples of each occluded and blurred character dataset are shown in Fig. \ref{fig:occluded_blurred}.

\section{More Experimental Results of Non-Zero-Shot Settings}
We conduct more experiments in non-zero-shot settings which represents that all categories in the testing dataset appear in the training dataset. For the non-zero-shot setting of handwritten characters, we use HWDB1.0-1.1 \cite{liu2013online} as the training dataset and ICDAR2013 \cite{yin2013icdar} as the testing dataset. For the non-zero-shot setting of scene characters, we use the characters collected from the training and testing dataset of CTW \cite{yuan2019large} as the training and testing dataset of this setting, respectively. In the non-zero-shot settings, We adopt the same training strategy as in the occluded and blurred settings, where the model is first pretrained by the radical-structured stroke tree representation and then fine-tuned by the character-level representation. As shown in Tab. \ref{tab:seen setting}, our method achieves the comparable performance in both the non-zero-shot settings of handwritten and scene characters. Please note that most of the previous methods utilize other strategies to achieve better performance (\textit{e.g.}, HDE \cite{cao2020zero} uses the data augmentation strategies and Template+Instance \cite{xiao2019template} adds additional printed characters as the prior knowledge).

\begin{table}[h]
\centering
\caption{More experimental results in the non-zero-shot settings of handwritten and scene characters. ``CharRep'' represents that we use the character-level representation to train or fine-tune our model. We achieve the comparable performance in the non-zero-shot settings.}
\scalebox{0.9}{
\begin{tabular}{c|cc}
\hline
Method & ICDAR2013 & CTW\\ 
\hline
Human Performance \cite{yin2013icdar} & 96.13\% & - \\
HCCR-GoogLeNet \cite{zhong2015high} & 96.35\% & - \\
ResNet152 \cite{he2016deep} & - & 80.94\% \\
DenseRAN \cite{wang2018denseran} & 96.66\% & 85.56\%  \\
FewshotRAN \cite{wang2019radical} & 96.97\% & 86.78\% \\
HDE \cite{cao2020zero} & 97.14\% & \textbf{89.25}\% \\
Chen et al. \cite{chen2021zero} & 96.74\% & 85.90\%  \\
Template+Instance \cite{xiao2019template} & \textbf{97.45}\% & - \\
\hline
CharRep & 96.17\% & 75.79\% \\
Ours & 96.05\% & 80.25\% \\
Ours + CharRep & 97.42\% & 86.28\% \\
\hline
\end{tabular}}

\label{tab:seen setting}
\end{table}

\section{Size of Confusable Character Set}
\begin{figure}[h]
    \centering
    \includegraphics[width=0.475\textwidth]{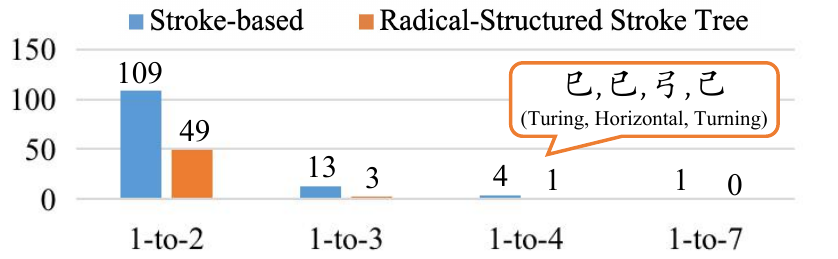}
    \caption{Compared with the stroke-based representation, the proposed RSST representation can mitigate the one-to-many problem.}
    \label{fig:one2many}
\end{figure}

As shown in Fig. \ref{fig:one2many}, compared with \cite{chen2021zero}, our method contains a smaller confusable set $\mathcal{C}$ since we take both radical structures and strokes into consideration for representing Chinese characters. Among 3,755 Level-1 characters, there are 280 confusable characters for the stroke-based approach and only 111 confusable characters for our radical-structured tree representation.

\section{Weighted Edit Distance for Radical-Structured Stroke Trees}
\begin{figure}[h]
    \centering
    \includegraphics[width=0.87\textwidth]{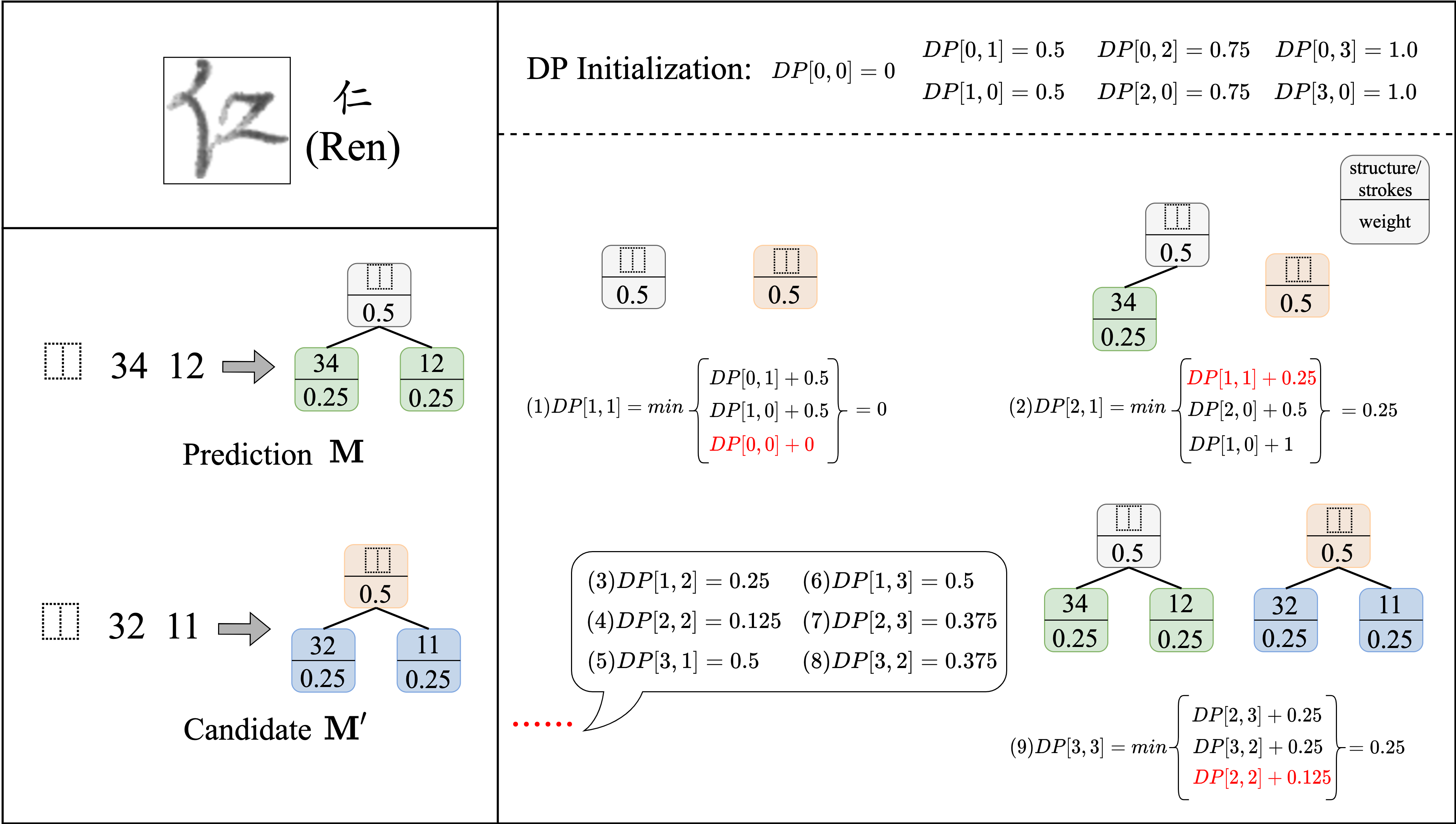}
    \caption{An example of calculating the Weight Edit Distance. $DP[3,3]$ is the Weight Edit Distance between the prediction $\mathbf{M}$ and the candidate $\mathbf{M}^\prime$}
    \label{fig:distance}
\end{figure}

In this section, we introduce the design of the proposed  Weighted Edit Distance. Inspired by HDE \cite{cao2020zero}, given the predicted radical-structured stroke tree, we assign a hierarchical attenuation weight to each node in the tree. Specifically, the weight of nodes in the $i$-th layer is $ \alpha^i$, where $\alpha$ is empirically set to 0.5. For the stroke nodes (\textit{i.e.}, the leaf nodes), we evenly distribute the weight of nodes to each stroke. We define $DP[i,j]$ as the distance between the first $i$ elements of the predicted tree $\mathbf{M}$ and the first $j$ elements of the candidate tree $\mathbf{M}^{\prime}$ in the depth-first-search sequence. When calculating $DP[i,j]$, the smallest of the following three cases is selected as the final value: (1) The sum of $DP[i-1,j]$ and the weight of the $i$-th element in $\mathbf{M}$. (2) The sum of $DP[i,j-1]$ and the weight of the $j$-th element in $\mathbf{M}^{\prime}$. (3) The sum of $DP[i-1,j-1]$ and the weighted edit distance between the $i$-th element in $\mathbf{M}$ and the $j$-th element in $\mathbf{M}^{\prime}$, which is computed by:

\begin{equation}
\frac{ED(\mathbf{M}_i, \mathbf{M}^{\prime}_j)\times w^{\prime}_j}{Len(\mathbf{M}^{\prime}_j)}
\end{equation}
where $w^{\prime}_j$ is the weight of the $j$-th element of $\mathbf{M}^{\prime}$. $ED$ and $Len$ denote the vanilla edit distance and the length of sequence. Therefore, we define the state transition equation $DP$ for dynamic programming as follows:

\begin{equation}
DP[i, j]=\text{min} \left\{
\begin{array}{l}
DP[i-1, j]+w_i, \\
DP[i, j-1]+w^{\prime}_j, \\
DP[i-1, j-1]+\frac{ED(\mathbf{M}_i, \mathbf{
M}^{\prime}_j)\times w^{\prime}_j}{Len(\mathbf{M}^{\prime}_j)}
\end{array}
\right.
\label{3}
\end{equation}
where $w_i$ is the weight of the $i$-th element of $\mathbf{M}$. An example of calculating the Weight Edit Distance is shown in Fig. \ref{fig:distance}.

\end{document}